\documentclass[twocolumn,showpacs,
aps,superscriptaddress,
prd,notitlepage,showkeys,
nofootinbib, floatfix]{revtex4-1}

\usepackage[normalem]{ulem}
\usepackage{amssymb}
\usepackage{amsmath}
\usepackage{graphicx}
\usepackage{dcolumn}
\usepackage[colorlinks,urlcolor=blue,citecolor=blue,linkcolor=blue]{hyperref}
\usepackage{color,units}
\usepackage[dvipsnames]{xcolor}
\usepackage{lineno}
\usepackage{xspace}
\usepackage{orcidlink}
\usepackage{longtable}
\usepackage{float}
\usepackage{multirow}
\usepackage{amsfonts,wasysym,epsfig, verbatim, subfigure, bm,mathrsfs,lipsum}
\usepackage{bm}
\usepackage{soul}
\usepackage{pifont}
\usepackage{tikz}
\usepackage{booktabs}
\usetikzlibrary{shapes.geometric, arrows, shadows, positioning}
\usetikzlibrary{positioning, shapes.misc}

\usepackage{multirow}
\usepackage{longtable,tabu}

\definecolor{lime}{HTML}{A6CE39}
\DeclareRobustCommand{\orcidicon}{%
	\begin{tikzpicture}
	\draw[lime,fill=lime] (0,0) 
	circle [radius=0.16] 
	node[white] {{\fontfamily{qag}\selectfont \tiny ID}};
	\draw[white,fill=white] (-0.0625,0.095) 
	circle [radius=0.007];
	\end{tikzpicture}
	\hspace{-2mm}
}
\foreach \x in {A,...,Z}{%
	\expandafter\xdef\csname orcid\x\endcsname{\noexpand\href{https://orcid.org/\csname orcidauthor\x\endcsname}{\noexpand\orcidicon}}
}

\begin{document}
\title{Investigating the efficacy of vortex beam speckle patterns for the recognition of atmospheric turbulence and fractional topological charge with a deep learning model}
\author{Ratikanta Swain\orcidA}
\affiliation{Department of Physics, Indian Institute of Technology, Patna 801106, India}
\author{Allarakha Shikder\orcidB}
\affiliation{Department of Physics, Indian Institute of Technology, Patna 801106, India}
\author{Naseem Babu\orcidC}
\affiliation{Dept. of Artificial Intelligence and Machine Learning, Manipal University Jaipur, Rajasthan, 303007, India}
\author{Niladri Sarkar\orcidE}
\affiliation{Department of Physics, Indian Institute of Technology, Patna 801106, India}
\author{Aryan Raj\orcidF}
\affiliation{Department of Physics, Indian Institute of Technology, Patna 801106, India}
\author{Raghavan K. Easwaran\orcidD}
\email{raghavanke@iitp.ac.in}
\affiliation{Department of Physics, Indian Institute of Technology, Patna 801106, India}

\date{\today}

\begin{abstract}
Fractional topological charge (TC) enlarges the usable mode set of orbital angular momentum (OAM) links, but reading it reliably after atmospheric turbulence remains difficult. Interferometric and diffractive methods need an undistorted beam and a careful alignment, while most learned approaches treat integer or coarsely spaced charges and do not include turbulence. Here we ask whether the charge of a scalar vortex beam can be regressed directly from a single speckle image, and at which turbulence strength this stops working. We propagate vortex beams over 1500~m through 150 von K\'arm\'an phase screens, scatter them with a fixed random diffuser, and record $256\times256$, 16-bit speckle images for TC from 0 to 10 in steps of 0.01 and turbulence strengths $D/r_0$ from 0 to 50; the results reported here use a subset of 20\,000 images, and a data set of $10^{6}$ images is in preparation. A convolutional neural network estimates the TC and, through an auxiliary head, the turbulence category. On charges withheld from training, the mean absolute error is 0.031 without turbulence (with a common intensity scale) and rises to 0.78 at $D/r_0=10$ and 1.85 at $D/r_0=50$, against 2.5 for a constant prediction, and the turbulence category is identified correctly in 78\% of the images. The results show that, at the present training size, the charge is read accurately only without turbulence, and that the charge and the turbulence oth act on the speckle grain size, which may limit the readout. Experimental validation and the extension to vector vortex beams are left for future work.
\end{abstract}
\maketitle

\section{Introduction}
\label{intro}
Optical vortex beams carry a helical phase front, mathematically expressed as $\exp(i\ell\theta)$, where each photon of the beam carries an orbital angular momentum (OAM) of $\ell\hbar$ per photon\cite{allen1992}. This $\ell$ is also known as topological charge (TC) of the beam\cite{shen2019,balasubramaniam2026}. These beams underpin OAM-multiplexed communication, high-dimensional quantum information, optical trapping and super-resolution microscopy\cite{willner2015,willner2021,ng2010,tao2005,erhard2018}. When the azimuthal phase step is not a multiple of $2\pi$, the TC is fractional: the intensity ring acquires a radial dark notch, and the field is a coherent superposition of integer vortex modes\cite{berry2004,leach2004,gotte2008,zhang2022}. Fractional TC adds a continuous degree of freedom. For example, a resolution of $0.1$ already enlarges the usable mode set tenfold over integers\cite{zhang2022}. Exploiting it requires reading the charge accurately, because an error of even a fraction of a unit produces crosstalk between channels\cite{paterson2005,tyler2009,rodenburg2012,krenn2014}.\\
Traditional TC measurements rely on interferometric fork fringes or on diffraction through apertures and gratings\cite{hickmann2010,li2015,huang2012,alperin2016,hosseinisaber2020,zhu2016,daryabi2023,zheng2025,shikder2023}. For fractional charges the reported resolutions are $0.05$, $0.1$, $0.25$ and $0.5$, with errors of only a few per cent\cite{deng2019,shikder2024}. All of these studies assume an undistorted beam and a carefully aligned optical set-up, and their efficiency decreases once the beam is scattered or the wavefront is randomised\cite{lin2023}.\\
A fractional vortex beam comprises a superposition of integer vortex modes, meaning its OAM is defined by an OAM spectrum concentrated on the integers nearest to $\ell$. Because fractional charges break mode orthogonality, measuring this power spectrum is significantly harder than measuring the topological charge alone\cite{lin2023,derrico2017,berry2004}. A recent single-shot method overcomes this by pairing a single-path interferometer with a coaxial reference beam and polarisation phase-shifting to recover the phase and project it onto spiral harmonics, measuring spectra up to $\ell \approx 30$ with an accuracy better than 0.1 at $\ell = 4.5$ and $4.75$\cite{sarkar2024}.\\
Deep learning has loosened these constraints, but mostly as classification of integer charges\cite{liu2019,zhu2023,hu2023}. For fractional charges, a DenseNet was reported to reach a resolution of $0.01$ with $98\%$ accuracy on $10\,000$ images\cite{liu2019,huang2017}, and a DenseNet applied to the cross-spectral density\cite{wang2025} of partially coherent beams reached $99.99\%$ accuracy at $\Delta\ell = 0.01$, against $78\%$ and $92\%$ from intensity alone for the two beam families under low coherence. That work is simulated, covers one unit of charge ($\ell = 1.00$ to $1.99$), and uses the cross-spectral density, a second-order coherence quantity rather than a single intensity frame, and it was carried out without turbulence.\\
Atmospheric turbulence is the central obstacle outside the laboratory~\cite{paterson2005,krenn2014}. It distorts the helical wavefront and couples neighbouring OAM states. Cao et al.~\cite{cao2021} recognised fractional OAM through simulated turbulence with a residual network applied to two-mode ``petal'' beam intensities at $\Delta\ell = 0.1$; accuracy fell from $99.91\%$ to $94.97\%$ and $85.30\%$ as $C_n^2$ increased to $5\times 10^{-14}\ \mathrm{m^{-2/3}}$ over $1500\ \mathrm{m}$. That study used a purpose-designed hybrid beam, ten classes ($2.0$ to $2.9$) and the directly imaged beam intensity, so fine-grained fractional TC through turbulence from a generic vortex beam remains open.\\
Speckle offers a different route. A scattering medium randomises the field, yet the speckle keeps the statistics of its source: the grain size is set by the illuminated area, shrinks as the charge grows, and the autocorrelation becomes asymmetric for fractional charge. Hu et al.~\cite{hu2020} showed that the speckle size is fixed by the total illuminated area and not by the phase or amplitude structure of the beam, so the charge enters only because a larger charge widens the bright ring~\cite{hu2020}. A visual geometry group network with 16 weight layers (VGG-16)\cite{simonyan2015} applied to experimental speckle estimated the TC from $0$ to $10$ with a mean absolute error of $0.039$ when all charges were in training, and $0.12$ to $0.16$ for unseen intermediate charges\cite{mutupuri2026} when trained on integers or half-integers. That study used a single scattering geometry without turbulence, and its training and test patches were taken from the same recorded frame; the authors acknowledge that this limits the conclusions to generalisation across regions of a single acquisition. Learned recognition of structured light from speckle has also been shown more broadly\cite{li2018,hu2023,khare2024}.\\
Speckle also carries the properties of the medium. Texture statistics of vortex-beam speckle separated three $C_n^2$ levels more cleanly than Gaussian-beam speckle, using single instantaneous frames\cite{khare2024}. A network classified $20$ turbulence levels\cite{jiang2025} ($D/r_0$ from $10$ to $40$) from speckle intensity with about $68\%$ accuracy, rising to about $81\%$ when the OAM spectrum was added. Turbulence strength, the nuisance parameter that degrades charge readout, is therefore itself encoded in the same kind of image. This motivates our design: one network estimates the charge and, through an auxiliary classification head, the turbulence category, and its error is reported separately for each turbulence level.\\
These results leave a specific question open. Speckle size is set by the illuminated area, and turbulence itself broadens, wanders and breaks up the beam, so a single speckle frame may not separate a larger charge from stronger turbulent spreading. Whether the charge can still be recovered, and at which turbulence strength it can no longer be, has not been determined. The answer matters in practice, because the strongly turbulent regime is where free-space OAM links are hardest to operate.\\
To our knowledge, no study among those we reviewed regresses a fractional TC at a fine step from single-frame speckle of a scalar vortex beam that has crossed turbulence, or reports its error for each turbulence strength. Without such a result, it is not known whether speckle can serve as an interferometer-free charge readout in a turbulent link, nor at which turbulence strength it stops working. We address this gap over a wide range of turbulence, from none to $D/r_0 = 50$. \\
We make three contributions. First, we build a wave-optics pipeline that propagates scalar (homogeneously polarised) vortex beams over $1500\ \mathrm{m}$ through $150$ von K\'arm\'an phase screens (outer scale $10\ \mathrm{m}$, inner scale $20\ \mathrm{mm}$), with a wavelength of $633.2\ \mathrm{nm}$, a beam waist of $2\ \mathrm{cm}$ and a $0.2\ \mathrm{m}$ receiver, a geometry similar to that of Ref.~\cite{cao2021}. The beams are generated at $D/r_0 = 0, 10, 20, 30, 40$ and $50$, scattered by a fixed random diffuser of full $2\pi$ phase range, and recorded as $256\times256$, 16-bit speckle images without cropping. The pipeline can produce $10^{6}$ images covering TC from $0$ to $10$ in steps of $0.01$, each with an independent turbulence realisation; here we use $20\,000$ of them ($4000$ beams with five turbulence realisations each), and the full set is in preparation. For $D/r_0$ of $10$ to $50$ the plane-wave Rytov variance of $1.3$ to $19$ takes the data into the strong, saturated scintillation regime. Second, we train a convolutional regressor that estimates the TC directly, with an auxiliary head for the turbulence category. We evaluate it on charges withheld from training, report its error separately for every $D/r_0$ level, and compare it with a speckle-size baseline and with a VGG-16 regressor of the type used in Ref.~\cite{mutupuri2026}. Third, we analyse how the readout error grows with turbulence strength and what limits it, including whether the charge is confounded with turbulent beam spreading.\\
Without turbulence the charge is recovered with a mean absolute error of $0.031$ on charges never seen in training, and the error grows with $D/r_0$, to $0.78$ at $D/r_0=10$ and $1.85$ at $D/r_0=50$. The results define where speckle-based fractional-TC readout holds and where turbulence erases the information, which is the practical limit for deployment.\\

The motivation for this work follows from a chain of related gaps. A fractional charge enlarges the usable OAM mode set, but only if the charge can be read with an error well below the channel spacing; otherwise neighbouring channels interfere~\cite{paterson2005,tyler2009}. The conventional readouts, interferometric and diffractive, are the natural tools for this, yet they need an undistorted beam and a carefully aligned set-up, so their efficiency drops just where free-space links operate, once the beam is scattered or its wavefront is randomised~\cite{li2015,hosseinisaber2020,deng2019}. Learned methods could relax these constraints, but the reported networks address integer or coarsely spaced charges, purpose-designed beams, or conditions without turbulence~\cite{cao2021,wang2025,mutupuri2026}, and to our knowledge none regresses a fine-step fractional charge of a generic vortex beam after turbulence. Speckle is an attractive way to fill this gap: a single frame needs no reference beam or interferometer, and its grain statistics carry both the charge and the strength of the turbulence~\cite{hu2020,khare2024,jiang2025}. What remains unknown is the practical limit, that is, whether the charge can still be recovered and at which turbulence strength it cannot, and this limit decides whether speckle-based readout can be deployed at all. A controlled simulation with known ground truth can establish it before any experiment is built, which is the aim of this work.

\section{Theoretical Model}
\label{TM}
We model the full process, from beam generation to charge estimation: a scalar vortex beam of real-valued topological charge $\ell$ is generated, propagates through a turbulent path, falls on a random diffuser, and the resulting speckle intensity is recorded and used to estimate $\ell$. Each step is described below with the equations used in the simulation.\\
\subsection{Scalar vortex beam with fractional charge}
The generated scalar vortex beam in the plane $z=0$, written in polar coordinates $(\rho,\varphi)$ as
\begin{align}
U_0(\rho,\varphi) &= A \Bigl(\frac{\rho}{w_0}\Bigr)^{|\ell|} \exp\!\Bigl(-\frac{\rho^{2}}{w_0^{2}}\Bigr) \nonumber \\ &\quad \times \exp(\mathrm{i}\ell\varphi) , 
\label{eq:beam} 
\end{align}
with beam waist $w_0=2\ \mathrm{cm}$ and $\ell\in[0,10]$ real. For integer $\ell$ this is the $p=0$ Laguerre-Gaussian mode at its waist. For non-integer $\ell$ the azimuthal phase does not close on itself: the phases at $\varphi=\pi$ and $\varphi=-\pi$ differ by $2\pi\ell$, which is a multiple of $2\pi$ only for integer $\ell$, so the phase is discontinuous across the cut and the field is not an OAM eigenstate but a superposition of integer OAM modes, as shown next. Expanding the azimuthal factor into integer OAM components, all sharing the same radial profile,
\begin{align}
    \exp(\mathrm{i}\ell\varphi) &= \sum_{m=-\infty}^{\infty} c_m \exp(\mathrm{i} m\varphi) , \nonumber \\
    c_m &= \frac{1}{2\pi} \int_{-\pi}^{\pi} \mathrm{e}^{\mathrm{i}(\ell-m)\varphi} \, \mathrm{d}\varphi = \operatorname{sinc}(\ell-m) , \label{eq:berry}
\end{align}

with $\operatorname{sinc}(x)=\sin(\pi x)/(\pi x)$. Since the radial factor is common to all components, the power in the azimuthal component $m$ is therefore $P_m=\operatorname{sinc}^{2}(\ell-m)$, which sums to one and is concentrated mainly on the two integers nearest to $\ell$. A fractional charge is thus a superposition of integer modes, and any process that couples neighbouring modes, as turbulence does, changes this spectrum. The overall amplitude $A$ is fixed by making every beam with unit power, so only the shape of the field matters.
\subsection{Von K\'arm\'an turbulence and phase screens}
Atmospheric turbulence is the random wind-driven mixing of air of different temperature which makes the refractive index of the air fluctuate randomly in space and time. A beam crossing such a medium accumulates random phase distortions, so its wavefront is distorted, its intensity scintillates, and the beam wanders and spreads~\cite{andrews2005,tatarskii1961}. The statistics of these fluctuations are classically described by the Kolmogorov spectrum~\cite{kolmogorov1941}, which diverges at low spatial frequency. The von K\'arm\'an spectrum~\cite{karman1948,andrews2005} removes this by introducing an outer and an inner scale, so it stays finite at very low and very high spatial frequencies and we use it to generate the turbulence conditions of the study. The refractive-index fluctuations along the path are represented by $N_s=150$ thin random phase screens, separated by $\Delta z=L/N_s=10\ \mathrm{m}$ over the total path $L=1500\ \mathrm{m}$ (wavelength $\lambda=633.2\ \mathrm{nm}$, wavenumber $k=2\pi/\lambda$). Each screen is a Gaussian random field with the von K\'arm\'an power spectral density of the phase, in cycles per metre,
\begin{align}
    \Phi_\phi(f) &= 0.023 \, r_0^{-5/3} \bigl(f^{2}+f_0^{2}\bigr)^{-11/6} \nonumber \\
    &\quad \times \exp\!\Bigl[-\Bigl(\frac{f}{f_m}\Bigr)^{2}\Bigr] , \nonumber \\
    f_0 &= \frac{1}{L_0} , \qquad f_m = \frac{5.92}{2\pi\,l_0} , \label{eq:psd}
\end{align}
where $r_0$ is the Fried parameter of the screen, $L_0=10\ \mathrm{m}$ is the outer scale and $l_0$ the inner scale. We set $l_0=L_0/\mathrm{Re}^{3/4}=20\ \mathrm{mm}$ with the Reynolds number $\mathrm{Re}=3960$, which lies within the range of Reynolds numbers used in \cite{jiang2025}. The outer scale removes the divergence of the Kolmogorov spectrum at low frequency, and the inner scale cuts off the spectrum at high frequency.\\
The total turbulence is quantified by the ratio $D/r_0$, where $D=0.2\ \mathrm{m}$ is the width of the receiver window and $r_0$ is the Fried parameter of the whole path. For a plane wave, $r_0=(0.423\,k^{2}C_n^{2}L)^{-3/5}$ with refractive-index structure constant $C_n^{2}$, and the Rytov variance is $\sigma_R^{2}=1.23\,C_n^{2}k^{7/6}L^{11/6}$. Because the phase variances of independent screens add and the spectrum scales as $r_0^{-5/3}$, $N_s$ equal screens reproduce the total $r_0$ when each screen has
\begin{equation}
r_{0,s}=r_0\,N_s^{3/5}.
\label{eq:r0s}
\end{equation}
The levels of the study are listed in Table~\ref{tab:levels}. For $D/r_0\geq10$ the Rytov variance exceeds one, so the beam is in the moderate-to-strong and saturated scintillation regime, where weak-turbulence theory does not apply and the wave-optics simulation is the reference. $D/r_0=0$ means no turbulence.

\begin{table}[h] 
\centering 
\caption{Turbulence levels. $r_0$ is the plane-wave Fried parameter for $D=0.2\ \mathrm{m}$ over $1500\ \mathrm{m}$; $C_n^{2}$ and the Rytov variance $\sigma_R^{2}$ follow from the formulas in the text; $r_{0,s}$ is the Fried parameter of each of the 150 screens (Eq.~\ref{eq:r0s}).} 
\label{tab:levels} 
\begin{tabular}{rrrrr} 
\toprule
$D/r_0$ & $r_0$ (mm) & $C_n^{2}$ ($\mathrm{m^{-2/3}}$) & $\sigma_R^{2}$ & $r_{0,s}$ (mm)\\ 
\midrule
10 & 20.0 & $1.09\times10^{-14}$ & 1.29 & 404\\ 
20 & 10.0 & $3.45\times10^{-14}$ & 4.10 & 202\\ 
30 & 6.7  & $6.78\times10^{-14}$ & 8.06 & 135\\ 
40 & 5.0  & $1.10\times10^{-13}$ & 13.03 & 101\\ 
50 & 4.0  & $1.59\times10^{-13}$ & 18.89 & 81\\ 
\bottomrule
\end{tabular}
\end{table}

Each screen on the $M\times M$ grid ($M=512$, pixel pitch $\delta=0.78125\ \mathrm{mm}$, grid width $0.4\ \mathrm{m}$) is generated by the spectral method,
\begin{align}
\phi(\mathbf{x}) &= \mathcal{F}^{-1}\Bigl\{\bigl(a+\mathrm{i} b\bigr)\sqrt{\Phi_\phi(\mathbf f)}\,\Delta f\Bigr\} , \nonumber \\ a,b &\sim \mathcal{N}(0,1) , \qquad \Delta f = \frac{1}{M\delta} , 
\label{eq:screen} 
\end{align}
where the real and imaginary parts of one transform give two independent screens. A discrete grid cannot hold frequencies below $\Delta f$, which carry most of the low-order phase variance. We therefore add subharmonics at three levels, with frequencies $\Delta f/3^{p}$, $p=1,2,3$, on a $3\times3$ lattice each. The simulated structure function $D_\phi(r)=\langle[\phi(\mathbf x)-\phi(\mathbf x+\mathbf r)]^{2}\rangle$ is compared with
\begin{align}
D_\phi(r) &= 4\pi \int_0^{\infty} \bigl[1-J_0(2\pi f r)\bigr] \nonumber \\ &\qquad \times \Phi_\phi(f) \, f \, \mathrm{d} f , 
\label{eq:dphi} 
\end{align}
in Sec.~\ref{sec:checks}.
\subsection{Propagation through the turbulent path} 
Each step applies one phase screen and then free-space propagation over $\Delta z$ in the paraxial (Fresnel) approximation, the split-step angular-spectrum method:
\begin{align}
U_{j+1} &= \mathcal{F}^{-1}\Bigl\{H(\mathbf f)\,\mathcal{F}\bigl\{U_j\,\mathrm{e}^{\mathrm{i}\phi_j}\bigr\}\Bigr\} , \nonumber \\ H(\mathbf f) &= \exp\!\bigl(-\mathrm{i}\pi\lambda\,\Delta z\,|\mathbf f|^{2}\bigr) , \quad j=0,\dots,N_s-1 . 
\label{eq:split} 
\end{align}
Without turbulence the same propagation is done in one step over $L$. The simulation grid is $0.4\ \mathrm{m}$ wide, and the receiver records only its central $0.2\ \mathrm{m}\times0.2\ \mathrm{m}$ window ($256\times256$ pixels), $U_{\mathrm{rx}}=U_{N_s}\big|_{\mathrm{window}}$. The larger grid prevents light that is scattered out of the window from re-entering through the periodic boundary of the Fourier transform: on a grid of only $0.2\ \mathrm{m}$ the wrapped power reaches 8-16\% at $D/r_0=30$-$50$, whereas on the $0.4\ \mathrm{m}$ grid the power at its edge stays at or below about $0.02\%$. The beam intensity before the diffuser,
\begin{equation}
I_b(\mathbf x)=|U_{\mathrm{rx}}(\mathbf x)|^{2}, 
\label{eq:ib} 
\end{equation}
shows the effect of turbulence directly as beam wander (shift of the centroid), broadening (growth of the rms radius) and break-up of the ring. At $\ell=10$ the fraction of the power inside the window falls from $0.998$ at $D/r_0=10$ to $0.829$ at $D/r_0=50$.

\subsection{Diffuser and speckle formation}
The receiver field illuminates a thin random phase diffuser with transmission
\begin{align}
t(\mathbf x) &= \exp\!\bigl[\mathrm{i}\psi(\mathbf x)\bigr] , \qquad \psi\ \text{independent in every pixel} , \nonumber \\ \psi &\sim \mathcal{U}[0,\,2\pi a] , \quad a=1 , 
\label{eq:diffuser} 
\end{align}
drawn once from a fixed seed and used for all samples. The mean transmission is $\langle t\rangle=\mathrm{e}^{\mathrm{i}\pi a}\operatorname{sinc}(a)$, so the fraction of light that passes unscattered is $\operatorname{sinc}^{2}(a)$. For $a=1$ it vanishes and the speckle is fully developed. The speckle is observed in the far field of the diffuser (Fraunhofer regime), where the intensity is the squared modulus of the Fourier transform of the transmitted field,
\begin{equation} 
I_s(\mathbf f) = \Bigl|\mathcal{F}\bigl\{U_{\mathrm{rx}}(\mathbf x)\,t(\mathbf x)\bigr\}(\mathbf f)\Bigr|^{2} . 
\label{eq:speckle} 
\end{equation}
The field is zero-padded to $4\times$ its width before the transform. The padding factor $P=4$ interpolates the far-field pattern $P$ times finer, which plays the role of the diffuser-camera distance $z$ in a real set-up: the speckle grain, measured as the full width at half maximum of the intensity autocorrelation, is about $16\ \mathrm{px}$ at $\ell=0$ and $4.7\ \mathrm{px}$ at $\ell=10$ for a beam without turbulence, a ratio of $3.4$. The central $256\times256$ pixels are kept and stored as 16-bit integers on a common intensity scale, $I_{16}=\mathrm{round}\big[65535\,I_s/32\big]$, with every beam launched with unit power.\\
For a diffuser that is uncorrelated from pixel to pixel and has a full-range phase, the speckle field is a circular complex Gaussian variable. The intensity then follows negative-exponential statistics with contrast $C=\sigma_I/\langle I\rangle=1$, and a fraction $1-\mathrm{e}^{-1}=0.632$ of the pixels lie below the mean. The second-order correlation is given by the Siegert relation,
\begin{align}
\langle I_s(\mathbf f)\,I_s(\mathbf f+\Delta\mathbf f)\rangle &= \langle I_s\rangle^{2}\bigl[1+|\mu(\Delta\mathbf f)|^{2}\bigr] , \nonumber \\ \mu(\Delta\mathbf f) &= \frac{\int I_b(\mathbf x)\,\mathrm{e}^{-2\pi\mathrm{i}\,\mathbf x\cdot\Delta\mathbf f}\,\mathrm{d}^{2}x}{\int I_b(\mathbf x)\,\mathrm{d}^{2}x} , \
\label{eq:siegert} 
\end{align}
so the grain structure is set by the Fourier transform of the intensity $I_b$ \emph{before} the diffuser (van Cittert-Zernike). The width of $|\mu|^{2}$ scales as the inverse of the rms radius $\rho_{\mathrm{rms}}$ of $I_b$. For the beam of Eq.~\eqref{eq:beam} in free space, $I_b\propto\rho^{2|\ell|}\exp(-2\rho^{2}/w_0^{2})$ and
\begin{align}
\rho_{\mathrm{rms}}^{2} &= \frac{\int\rho^{2}I_b\,\mathrm{d}^{2}x}{\int I_b\,\mathrm{d}^{2}x} = \frac{w_0^{2}}{2}\bigl(|\ell|+1\bigr) , \nonumber \\ \Delta s &\propto \frac{1}{\rho_{\mathrm{rms}}} \propto \frac{1}{\sqrt{|\ell|+1}} , 
\label{eq:size} 
\end{align}
which is the scaling reported for the speckle size of Laguerre-Gaussian beams\cite{hu2020}. It predicts a grain-size ratio of $\sqrt{11}\approx3.3$ between $\ell=0$ and $\ell=10$. Two consequences follow for turbulent beams. First, a translation of $I_b$ (beam wander) changes only the phase of $\mu$, not $|\mu|$, so wander is invisible in the second-order statistics. Second, broadening and break-up of the beam change $\rho_{\mathrm{rms}}$ and the shape of $|\mu|$ in the same way as a change of $\ell$ does, so the grain size alone cannot separate the two. What can separate them is the \emph{shape} of $|\mu|^{2}$, since a ring and a filled spot of the same rms radius have different autocorrelations. For a fixed diffuser, moreover, $I_s$ is a deterministic function of the complex field $U_{\mathrm{rx}}$ and carries information beyond second-order statistics. Because one diffuser is used in the whole dataset, a learned estimator may exploit it and we have not yet tested other diffusers.

\subsection{Estimation problem} 
Given one speckle image $I_s$ of an unknown charge $\ell\in[0,10]$ and unknown turbulence $D/r_0$, the task is to find an estimate $\hat\ell=f_\theta(I_s)$. The estimator $f_\theta$ is a convolutional network with parameters $\theta$ and two outputs, a regression of $y=\ell/10$ and a classification of the turbulence category $c\in\{$none, weak, moderate, strong$\}$ ($D/r_0=0$; $10,20$; $30$; $40,50$; these are labels for the classifier only, since every level with $D/r_0\geq10$ has $\sigma_R^{2}>1$).\\
Training minimises 
\begin{align}
\mathcal{L}(\theta) &= \frac{1}{B}\sum_{b=1}^{B}\Bigl[\mathrm{SL1}_{\beta}\bigl(\hat y_b-y_b\bigr) \nonumber \\ &\qquad - \lambda\sum_{k}\mathbb{1}[c_b=k]\log p_{b,k}\Bigr] , \nonumber \\ \lambda &= 0.1 , \quad \beta = 0.02 , 
\label{eq:loss} 
\end{align} where $\mathrm{SL1}_\beta(e)=e^{2}/(2\beta)$ for $|e|<\beta$ and $|e|-\beta/2$ otherwise (smooth $L_1$), and $p_{b,k}$ are the softmax probabilities of the category head.We report the mean absolute error $\mathrm{MAE}=\langle|\hat\ell-\ell|\rangle$, always separately for each $D/r_0$ level. To measure generalisation to charges the network has never seen, all charges with $\mathrm{round}(100\,\ell)\bmod10=5$ ($\ell=0.05,0.15,\dots,9.95$) are excluded from training, and the validation and test sets contain only these charges. The network is a residual convolutional network with $4.06$ million parameters, trained for $20$ epochs with AdamW (learning rate $2\times10^{-3}$, weight decay $10^{-4}$) and a one-cycle schedule.

\subsection{Numerical consistency checks} 
\label{sec:checks} 
The following properties of the simulation were verified before the data set was generated. 
In free space the power after propagation equals the power before to within $10^{-6}$, and with turbulence the total power on the $0.4\ \mathrm{m}$ grid is conserved to $3\times10^{-5}$. The simulated structure function of the screens agrees with the exact discrete spectrum to within $0.8\%$ and, once subharmonics are included, with the continuum von K\'arm\'an structure function of Eq.~\eqref{eq:dphi} to within $4\%$ up to lags of $12.5\ \mathrm{mm}$, $6\%$ at $50\ \mathrm{mm}$ and $8.4\%$ at $100\ \mathrm{mm}$. For the speckle, the contrast is $0.97$ to $1.00$ and $62$ to $63\%$ of the pixels lie below the mean, as expected for fully developed speckle, and the speckle power spectrum agrees with the autocorrelation of $I_b$ predicted by Eq.~\eqref{eq:siegert} (log-correlation $0.98$--$0.995$ with its own beam, against $0.72$--$0.95$ with another beam; this check was made with an earlier padding factor of the simulation). Finally, every sample is a fixed function of its index: a sample generated alone is identical, to $2\times10^{-6}$, to the same sample generated inside a batch, and the graphics-processor implementation agrees with the CPU reference to $10^{-5}$ for identical phase screens.

\section{Results and Discussion}
\label{RnD}
In this section, we have presented the results obtained for the speckle images generated with the pipeline of Sec.~II. In Figs.~\ref{fig:gallery} and \ref{fig:gallery2}, we have shown the beam and its speckle for different turbulence strengths and different topological charges. In Fig.~\ref{fig:pred} and Table~\ref{tab:res}, the charge predicted by the network has been compared with the true charge for every $D/r_0$ level, and in Fig.~\ref{fig:mae} the error has been plotted against the turbulence strength. The turbulence category read from the same images is shown in Fig.~\ref{fig:conf}. Lastly, in Fig.~\ref{fig:train} we have looked at how the training proceeds, and in the text we have compared different ways of presenting the intensities to the network.
\subsection{Data used and evaluation protocol} \label{sec:protocol}
The results below come from $4000$ beams with five independent turbulence realisations each, that is, $20\,000$ images in total, generated with the pipeline of Sec.~II. The five images of a beam share the charge, the $D/r_0$ level and the diffuser, so the data are split by beam and never by image. As described in Sec.~II, all charges with $\mathrm{round}(100\,\ell)\bmod10=5$ are excluded from the training set, and the validation and test sets contain only these charges. The training set has $15\,080$ images, the validation set $1000$ and the test set $1265$ images, which come from $253$ beams; the remaining $2655$ images, which have ordinary charges but were assigned to the validation or test part of the original split, are not used. The network has been trained three times with different random seeds, and we report the mean and the standard deviation over the seeds. A network that always predicts $\hat\ell=5$ has an MAE of $2.5$ on these charges, and we use this number as the reference for a readout that has learned nothing.
\subsection{Beam and speckle images} 
\label{sec:gallery}
\begin{figure*}
    \centering 
    \includegraphics[width=\textwidth]{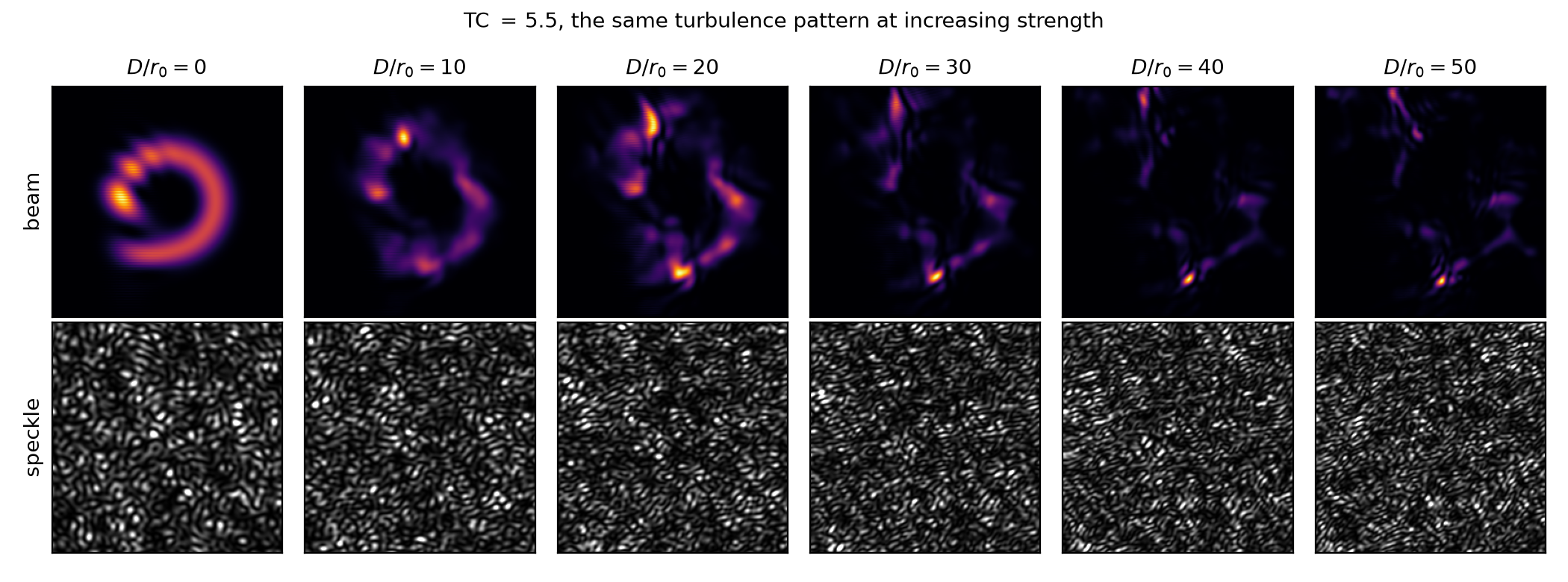} 
    \caption{Beam in the receiver window (top) and speckle (bottom) of a vortex beam with TC $=5.5$ for $D/r_0=0$, $10$, $20$, $30$, $40$ and $50$ (left to right). The same random turbulence pattern is used in every column. The beam is scaled to its own maximum.} 
    \label{fig:gallery} 
\end{figure*} 

\begin{figure*} 
    \centering 
    \includegraphics[width=0.8\textwidth]{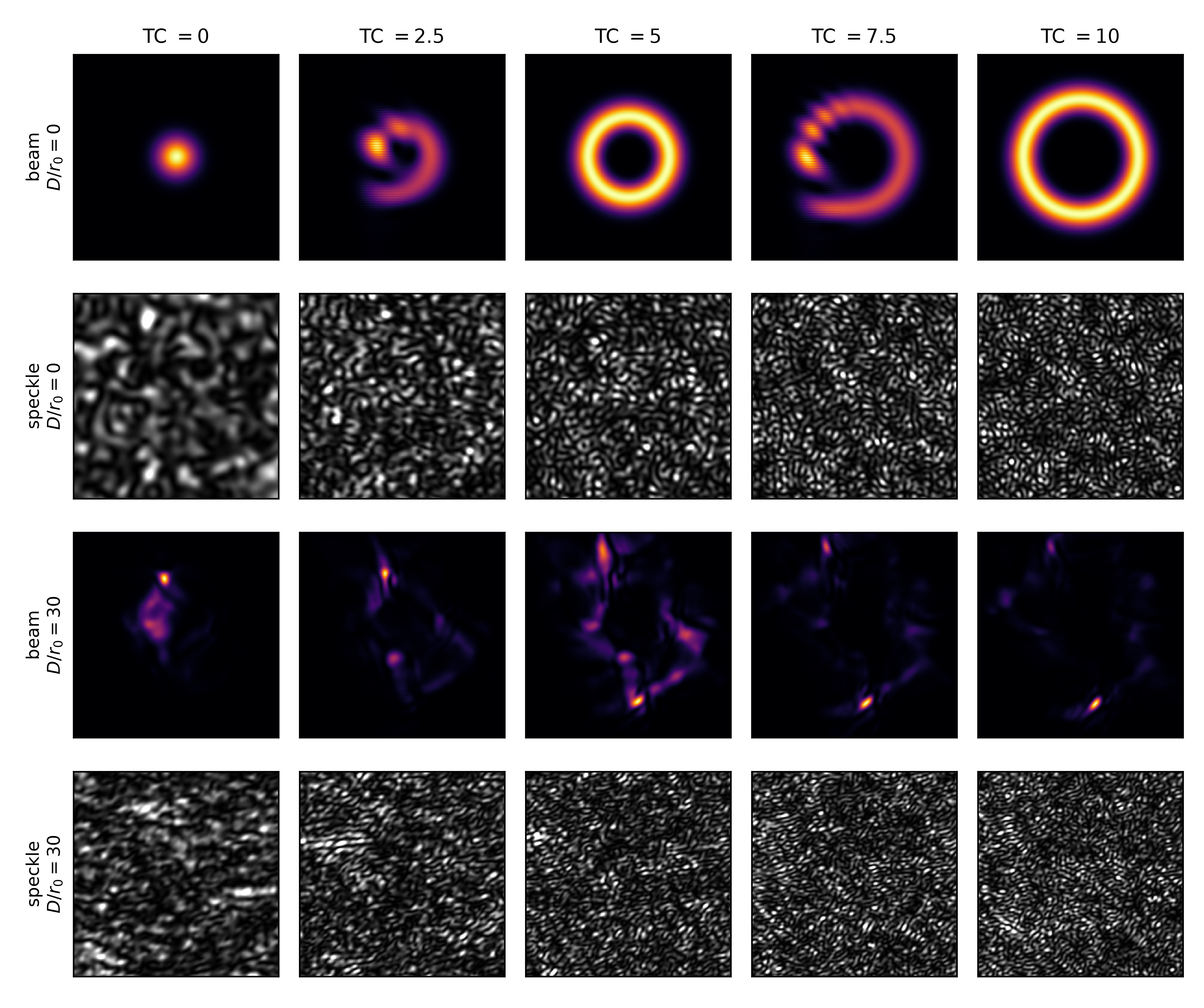} 
    \caption{Beam and speckle for TC $=0$, $2.5$, $5$, $7.5$ and $10$ (left to right) without turbulence ($D/r_0=0$, top two rows) and at $D/r_0=30$ (bottom two rows).} 
    \label{fig:gallery2} 
\end{figure*} 

Fig.~\ref{fig:gallery} shows how the same beam is changed as the turbulence becomes stronger, together with the speckle it produces behind the diffuser. Without turbulence, the beam is a ring with the radial notch of a fractional charge. Already at $D/r_0=10$ the ring is displaced and lumpy, and from $D/r_0=30$ onwards it is broken into a few bright filaments that carry most of the power. The speckle changes much less. Its grain stays fully developed: in a random subset of $600$ images the contrast is $0.99$ to $1.00$ at all $D/r_0$. The mean grain size falls from about $7.0$ to $5.7$ pixels as $D/r_0$ goes from $0$ to $50$, because turbulence broadens the beam, and a wider illuminated area gives a finer grain. The mean intensity in the window falls from $1.05$ to $0.90$ in units of the common intensity scale, and at $D/r_0=50$ it falls from $0.97$ for $\ell<2.5$ to $0.82$ for $\ell>7.5$, since turbulence scatters light out of the window and a larger charge has a wider ring.\\
Fig.~\ref{fig:gallery2} shows the same for five charges. Without turbulence, the grain shrinks clearly as the charge grows, which is the dependence on which the readout rests. At $D/r_0=30$ the speckle of the different charges is much harder to tell apart by eye. This is the main difficulty of the problem: a larger charge and a stronger turbulence both widen the beam and make the grain finer, so the charge and the turbulence act on the same grain statistics, as already discussed in Sec.~II.

\subsection{Charge regression} 
\label{sec:regression}
\begin{figure*} 
    \centering 
    \includegraphics[width=\textwidth]{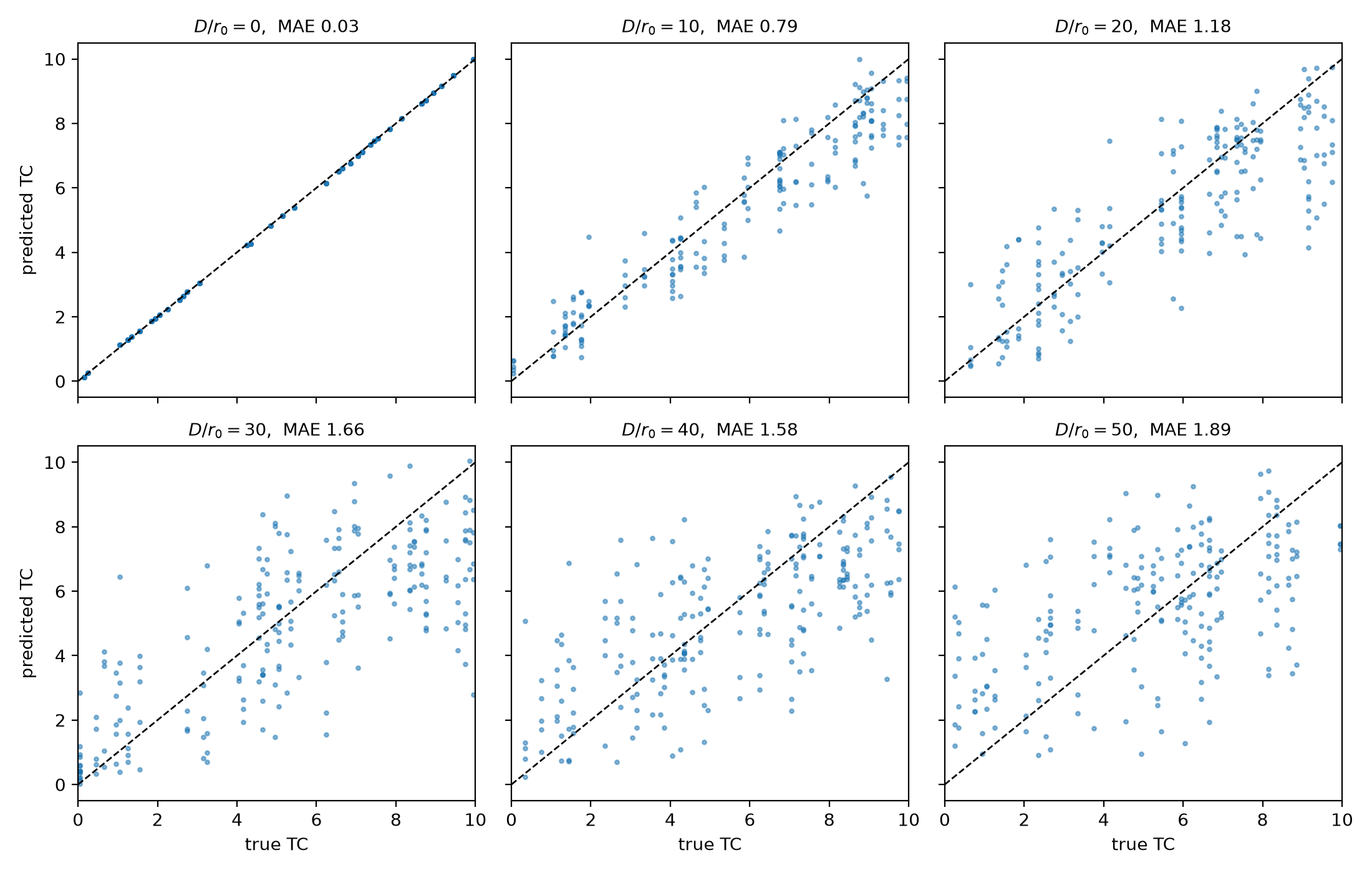} 
    \caption{Predicted against true TC for charges excluded from training, one panel per $D/r_0$ level (one seed; the MAE in each title refers to this seed). Each vertical column of points is one beam observed with five independent turbulence realisations, and these five points coincide at $D/r_0=0$. The dashed line is the ideal readout.} 
    \label{fig:pred} 
\end{figure*} 

\begin{table}
    \centering 
    \caption{MAE of the TC readout for each $D/r_0$ level on charges excluded from training (mean $\pm$ standard deviation over three seeds). The constant prediction has MAE $=2.5$.}
    \label{tab:res} 
    \begin{tabular}{rcc}\toprule
    $D/r_0$ & MAE & Reduction \\ \midrule
    0 & $0.031\pm0.008$ & $99\%$ \\
    10 & $0.781\pm0.018$ & $69\%$ \\
    20 & $1.217\pm0.041$ & $51\%$ \\
    30 & $1.622\pm0.037$ & $35\%$ \\
    40 & $1.599\pm0.047$ & $36\%$ \\
    50 & $1.850\pm0.068$ & $26\%$ \\
    all & $1.193\pm0.024$ & $52\%$ \\ \bottomrule
    \end{tabular}
\end{table}

\begin{figure} 
    \centering 
    \includegraphics[width=\columnwidth]{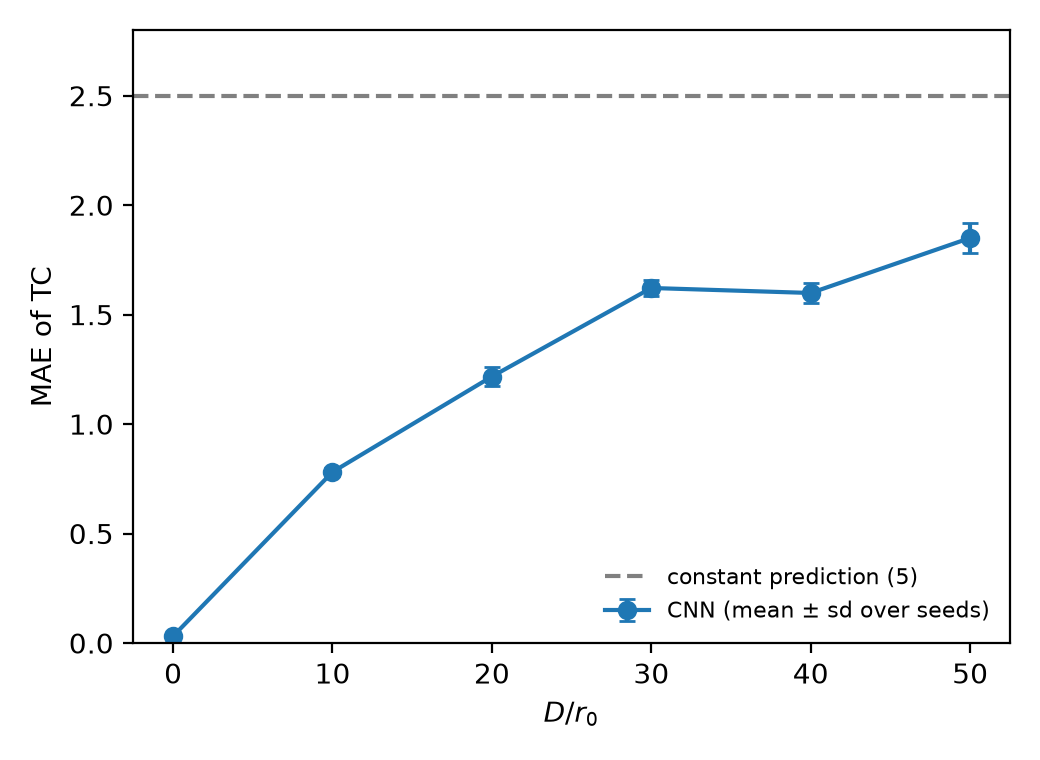} 
    \caption{MAE of the predicted TC against $D/r_0$ (mean and standard deviation over three seeds). The dashed line is the MAE of $2.5$ of a network that always predicts $\hat\ell=5$.} 
    \label{fig:mae} 
\end{figure} 

Fig.~\ref{fig:pred} compares the predicted and the true charge for each $D/r_0$, and Table~\ref{tab:res} lists the errors. Without turbulence, the network reads the charge very well: the points lie on the diagonal over the whole range from $0$ to $10$, and the MAE is $0.031\pm0.008$, for charges that the network has never seen during training.This value is obtained with the common intensity scale; with every image scaled to its own maximum it is $0.139$ (Sec.~\ref{sec:scale}). For comparison, Ref.~\cite{mutupuri2026} reports $0.12$ to $0.16$ for unseen intermediate charges ($0.039$ when all charges were in training) for experimental speckle without turbulence; the data sets differ, since ours is simulated and without detector noise, so the numbers are not directly comparable. As the diffuser is fixed, the speckle of a given charge is deterministic at $D/r_0=0$, and the result shows that the network interpolates between the charges it was trained on and does not only recognise them.\\
Turbulence changes this quickly. In Fig.~\ref{fig:pred}, the points spread around the diagonal already at $D/r_0=10$, and from $D/r_0=30$ the predictions are pulled toward the middle of the range: charges below $2$ are often predicted as $4$ to $6$, and charges above $8$ as $5$ to $8$. In Fig.~\ref{fig:mae}, we have shown how the MAE rises, to $0.781\pm0.018$ at $D/r_0=10$, $1.217\pm0.041$ at $20$, $1.622\pm0.037$ at $30$, $1.599\pm0.047$ at $40$ and $1.850\pm0.068$ at $50$, and it is $1.193\pm0.024$ over the whole test set. Measured against the constant prediction of $2.5$, the network removes $69\%$ of the error at $D/r_0=10$, $51\%$ at $20$, $35$ to $36\%$ at $30$ and $40$, and only $26\%$ at $50$. The increase is steep at first and then flattens, and the values at $30$ and $40$ cannot be told apart within the seed spread. The first turbulence step alone raises the error by a factor of about $25$. This means that, for the present training set, the speckle gives the charge to within a fraction of a unit only for weak turbulence. A decline of the accuracy with the turbulence strength has also been reported for the recognition of fractional charges from the beam intensity itself~\cite{cao2021}, and the present result shows that this decline persists, and is strong, for a fine-step readout from speckle.
\subsection{Turbulence category} 
\label{sec:category}
\begin{figure} 
    \centering 
    \includegraphics[width=0.85\columnwidth]{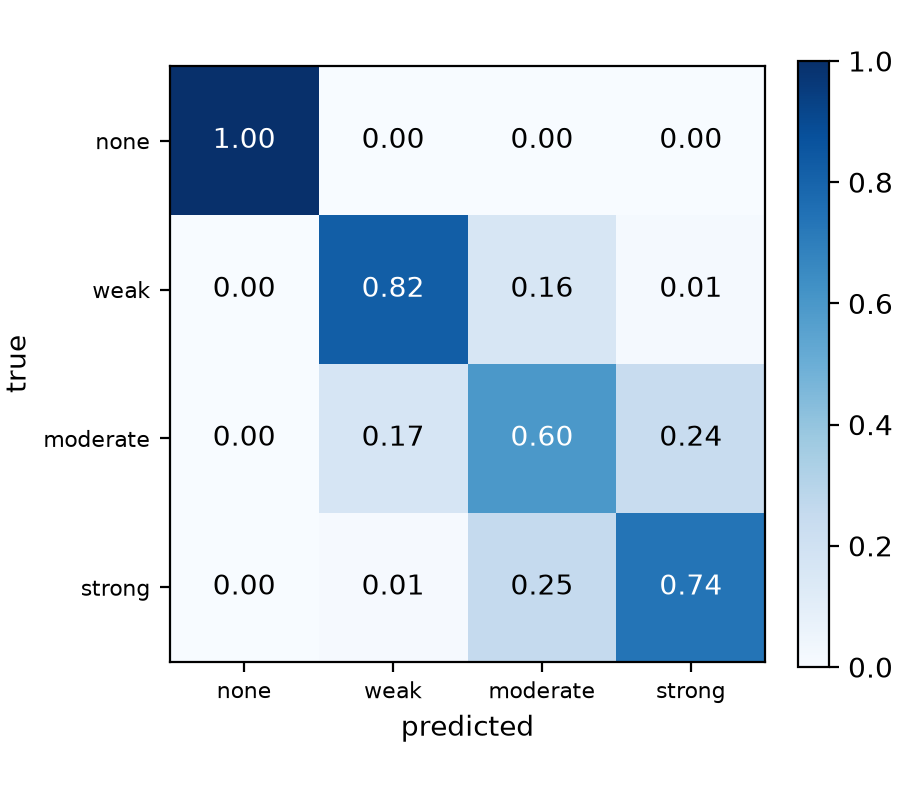} 
    \caption{Turbulence category predicted by the auxiliary head, row-normalised and summed over three seeds. The categories are none ($D/r_0=0$), weak ($10$ and $20$), moderate ($30$) and strong ($40$ and $50$).} 
    \label{fig:conf} 
\end{figure} 
The auxiliary head sorts the images into four categories, and Fig.~\ref{fig:conf} shows the resulting confusion matrix. The category accuracy is $0.783$ over the three seeds ($0.780$, $0.787$ and $0.782$). The absence of turbulence is always recognised, and the weak and strong categories are recognised in $82\%$ and $74\%$ of the cases. Moderate turbulence is the hardest, at $60\%$. The errors are almost all between neighbouring categories: moderate is taken for weak in $17\%$ and for strong in $24\%$ of the cases, while weak and strong are mistaken for each other in only about $1\%$ of the cases. This is what the speckle statistics lead us to expect. The contrast is close to $1$ at every level, so only the grain size, and to a smaller extent the mean intensity, carries information about the turbulence, and a single class lying between two others is bound to be confused with both.
\subsection{What limits the readout} 
\label{sec:limits} 
\begin{figure} 
    \centering 
    \includegraphics[width=\columnwidth]{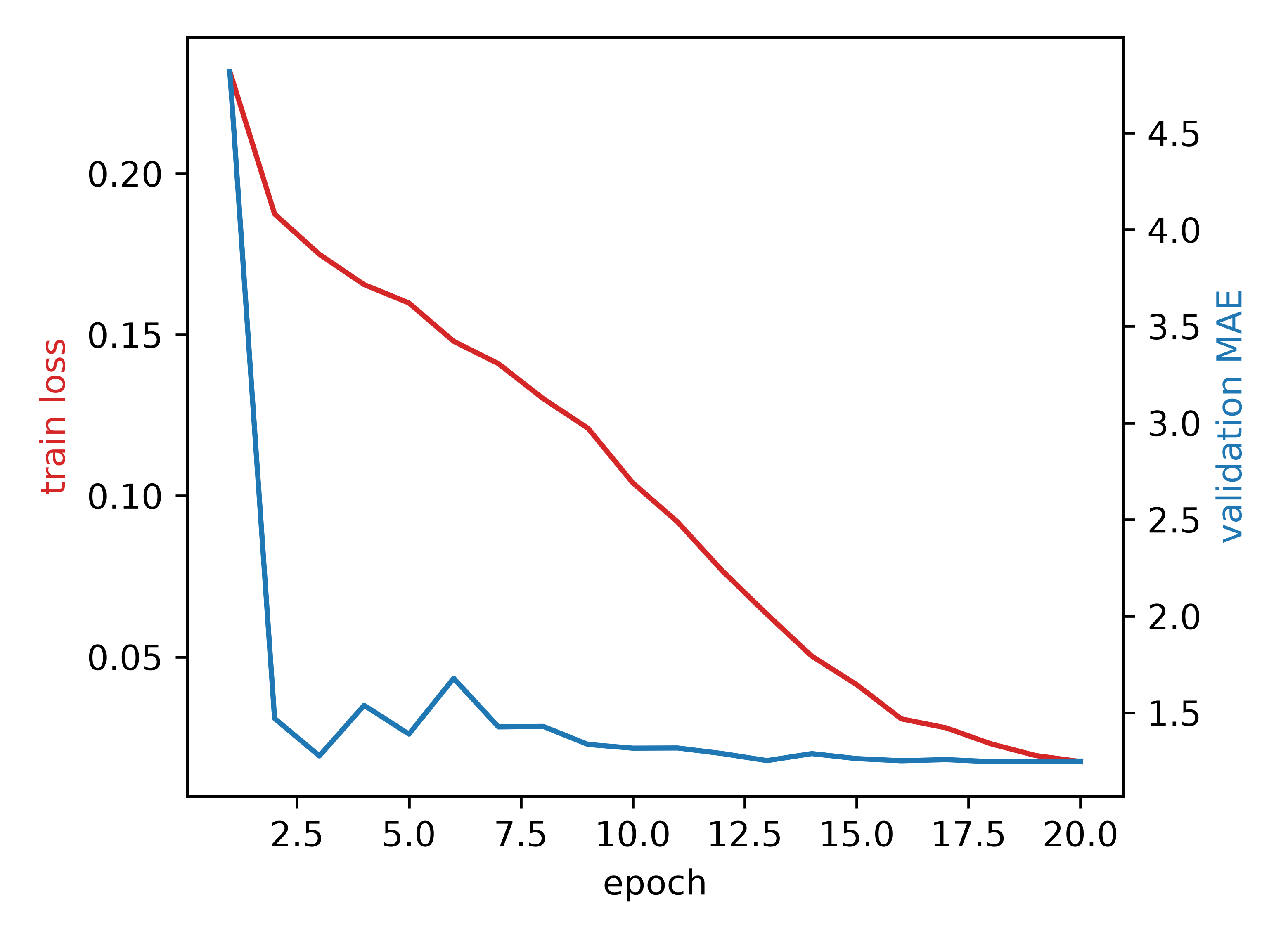} 
    \caption{Training loss (red, left axis) and validation MAE (blue, right axis) against the epoch for one seed.} 
    \label{fig:train} 
\end{figure} 
In Fig.~\ref{fig:train}, we have shown the training of one network. The validation error falls to about $1.3$ within three epochs and reaches $1.25$ at the end, while the training loss keeps decreasing from $0.23$ to $0.017$. The network therefore keeps learning the training images long after the validation error has stopped improving, which is a sign of memorisation. Memorisation is useful at $D/r_0=0$, where the speckle of a charge never changes, but it does not carry over to turbulent frames, since each of them is a new realisation. Together with the observation that the charge and the turbulence change the same grain statistics, this points to two possible causes of the large error at strong turbulence: too few turbulent realisations per charge for the network to separate the two effects, or a real ambiguity of the grain size that a single frame cannot resolve. The present data cannot decide between the two, and we regard this as the main open point of the work.
\subsection{Choice of intensity scale} 
\label{sec:scale}
We have also compared three ways of presenting the intensities to the network, each with three seeds. With every image scaled to its own maximum, the MAE is $1.253\pm0.033$. With the same images standardised by the mean and the standard deviation of the training set it is $1.223\pm0.042$, and with the common intensity scale used above (unit launch power for every beam) it is $1.193\pm0.024$. The first two cannot be told apart, which is expected since the first convolution is followed by batch normalisation. The common scale lowers the error at $D/r_0=0$ from $0.139$ to $0.031$ and makes the category accuracy more stable between the seeds ($0.78$ to $0.79$, against $0.53$ to $0.77$ for the other two), but the differences at the turbulent levels lie inside the seed spread. The common scale also lets the network use the absolute brightness, which carries the power scattered out of the window. This is real physics, but it requires a calibrated detector, so the per-image result is the safer reference for an experiment.
\subsection{Limitations}
\label{sec:limitations}
These results come from a simulation with a single fixed diffuser, one wavelength and scalar beams, and from $20\,000$ images. Other diffusers have not been tested, and the Rytov variance of $1.3$ to $19$ at $D/r_0=10$ to $50$ puts every turbulent level in the strong-scintillation regime. The numbers should therefore be read as the performance of this network at this training size and not as a limit set by the physics.

\section{Summary and Conclusions}
\label{SnC}
We set out to ask a simple question: can the fractional topological charge of a scalar vortex beam be read from a single speckle image after the beam has crossed atmospheric turbulence, and at which turbulence strength does this stop being possible? To address it, we propagated vortex beams over $1500\ \mathrm{m}$ through $150$ von K\'arm\'an phase screens, scattered them with a fixed random diffuser, and trained a convolutional network on the speckle images, with all test charges excluded from training.\\
The central finding is that the answer depends strongly on the turbulence. Without turbulence, the charge is recovered to a mean absolute error of $0.031$, even for charges that the network has never seen. With turbulence, the error rises quickly, to $0.78$ at $D/r_0=10$ and to $1.2$--$1.9$ at $D/r_0=20$ to $50$, compared with $2.5$ for a constant prediction, so that at strong turbulence the readout keeps only a weak trend. The turbulence category read from the same images is correct in $78\%$ of the cases, with almost all errors between neighbouring categories, and moderate turbulence is the hardest to identify. Standardising the inputs makes no difference, and a common intensity scale helps only at $D/r_0=0$ and for the category head, with the caveat that it requires a calibrated detector.\\
This outcome is not obvious a priori, because the turbulence strength is encoded in the same grain statistics as the charge. The turbulence broadens the beam and makes the grain finer in the same way as a larger charge does, and the images of Figs.~\ref{fig:gallery} and \ref{fig:gallery2} show that the speckle of different charges becomes hard to tell apart at strong turbulence even though the beam itself is strongly destroyed. The validation error stops improving after a few epochs while the training error keeps falling, so the present limit may well be set by the amount of turbulent data and by the use of a single frame, but our data cannot exclude a fundamental ambiguity of the grain size. In this sense, the result marks a first reference point for speckle-based charge readout in turbulence, and not its final limit.\\
These results are for a simulation with one fixed diffuser and scalar beams, and they come from a subset of the planned data. Training on the full set of $10^{6}$ images, using several frames of the same beam, adding spectral features of the speckle as input, and testing other diffusers will help us to understand how far this readout can be pushed. With an experimental check and the extension to vector vortex beams, the same approach may become a practical interferometer-free tool for reading fractional charges in turbulent optical links.

\begin{acknowledgments}
The authors thank colleagues at IIT Patna for useful discussions on optical vortex beams, atmospheric turbulence and speckle statistics.
\end{acknowledgments}

\makeatletter
\let\pre@bibdata\@empty
\makeatother
\bibliographystyle{apsrev4-1}
\bibliography{main}

\end{document}